\documentclass[sigconf]{acmart}
\usepackage{makecell}
\usepackage{balance}
\AtBeginDocument{%
  }

\copyrightyear{2024}
\acmYear{2024}
\setcopyright{acmlicensed}\acmConference[HRI '24 Companion]{Companion of the 2024 ACM/IEEE International Conference on Human-Robot Interaction}{March 11--14, 2024}{Boulder, CO, USA}
\acmBooktitle{Companion of the 2024 ACM/IEEE International Conference on Human-Robot Interaction (HRI '24 Companion), March 11--14, 2024, Boulder, CO, USA}
\acmDOI{10.1145/3610978.3640588}
\acmISBN{979-8-4007-0323-2/24/03}
\usepackage{fancyhdr} 
\usepackage{setspace} 
\begin{document}

\title{Unsupervised Motion Retargeting for Human-Robot Imitation}

\author{Louis Annabi}
\affiliation{%
 \institution{U2IS, ENSTA Paris, IP Paris}
 \city{Palaiseau}
 \country{France}
}
\email{louis.annabi@gmail.com}
\orcid{0000-0002-8645-0875}

\author{Ziqi Ma}
\affiliation{%
 \institution{U2IS, ENSTA Paris, IP Paris}
 \city{Palaiseau}
 \country{France}
}
\email{ziqi.ma@ensta-paris.fr}
\orcid{0009-0006-6775-4263}

\author{Sao Mai Nguyen}
\affiliation{%
 \institution{U2IS, ENSTA Paris, IP Paris}
 \institution{IMT Atlantique, Lab-STICC, UMR CNRS 6285}
% \city{Palaiseau}
 \country{France}
}
\email{nguyensmai@gmail.com}
\orcid{0000-0003-0929-0019}

% \author{Anonymous Author(s)}
% \affiliation{%
%   \institution{Institution}
%   \city{City}
%   \country{Country}
% }
% \email{email}

% \renewcommand{\shortauthors}{Annabi et al.}

\begin{abstract}

This early-stage research work aims to improve online human-robot imitation by translating sequences of joint positions from the domain of human motions to a domain of motions achievable by a given robot, thus constrained by its embodiment. Leveraging the generalization capabilities of deep learning methods, we address this problem by proposing an encoder-decoder neural network model performing domain-to-domain translation. In order to train such a model, one could use pairs of associated robot and human motions. Though, such paired data is extremely rare in practice, and tedious to collect. Therefore, we turn towards deep learning methods for unpaired domain-to-domain translation, that we adapt in order to perform human-robot imitation.

\end{abstract}






\maketitle
\thispagestyle{fancy}
\cfoot{
\begin{spacing}{0.9}
\scriptsize{
Annabi, L., Ma, Z., and Nguyen, S. M. (2024). Unsupervised Motion Retargeting for Human-Robot Imitation. Companion of the 2024 ACM/IEEE International Conference on Human-Robot Interaction(587--591). Association for Computing Machinery. \url{https://doi.org/10.1145/3610978.3640588}
 }
 \end{spacing}
}
%\vspace{20pt}}

\section{Introduction}

In many human-robot interaction scenarios, robots need to be able to imitate human motions \cite{Nehaniv2007,Billard2004RAS}. For example, imitating human motions can be used to reproduce human demonstrated motions \cite{Ijspeert2002RA2PIIIC,Schaal1997ANIPS}%in behavioral cloning \cite{Mandlekar2021}
, for coordination purposes in human-robot collaboration \cite{Wang2022}, or even to provide feedback to patients in physical rehabilitation scenarios \cite{Blanchard2022BRI}. This imitation is not a simple one-to-one mapping from human joint angle to motor angle, as the embodiment of humans and robots differ in sizes, proportions, velocities, forces, dynamics. Finding this relational mapping is referred to as the \textit{correspondence problem} \cite{Nehaniv2002IAA}. While the  correspondence problem between a single demonstrator and an imitator has been addressed in the human-robot interaction and the machine learning literature, the correspondence problem is the more acute when the robot needs to interact with different people. To our knowledge, a humanoid robot solving the correspondence problem of whole body movements from different demonstrators carrying out several tasks has not been addressed. While humanoid robots and different demonstrators can have the same skeletal structure as humans, their bone length and joint amplitudes vary from person to person. One difficulty in addressing this problem is the lack of complete dataset with all demonstrators carrying out the same set of tasks and synchronized and paired with a robot execution of the same set of tasks.  %Moreover, in this work, we consider an end-to-end approach from a video input.

\begin{figure}[ht]
    \centering
    \includegraphics[width=0.95\columnwidth]{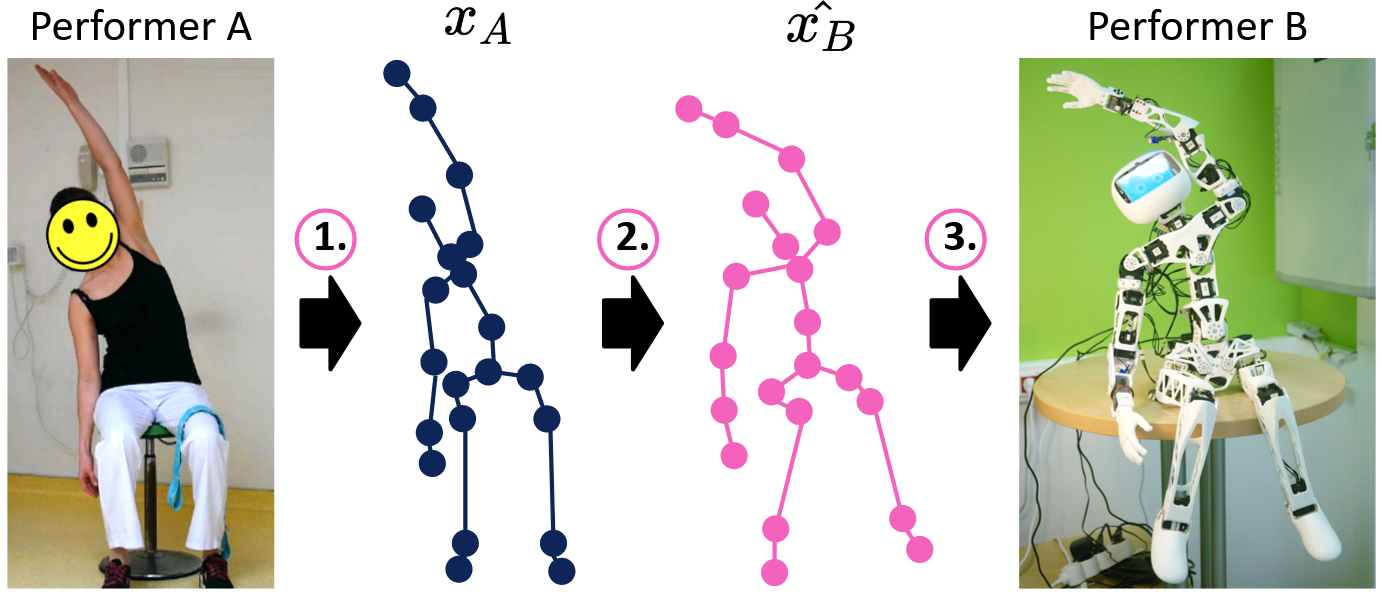}
    \vspace{-0.5em}
    \caption{The steps of the human-robot imitation process: (1) pose estimation outputs from a video from performer A a sequence of joint positions $x_A$, (2) motion retargeting translates joint positions $x_A$ into joint positions $x_B$ for performer B, (3) robot control  sends the low level control. }
    \label{fig:imitation_method}
        \vspace{-0.7em}
\end{figure}

We separate the motion imitation process in three steps: pose estimation, motion retargeting, and robot control, as represented in figure \ref{fig:imitation_method}. Pose estimation algorithms predict a sequence of skeleton joint positions from the human demonstrator given a sensor input. The motion retargeting step translates this sequence of joint positions towards a domain of joint positions achievable by the robot, i.e. from the  human embodiment to the robot embodiment spaces. Finally, one can use these sequences of joint positions as targets for motor control (e.g. dynamic movement). While there is a flourishing literature on pose estimation, few works have addressed the question of motion retargeting for human-robot imitation. In this work, we will examine the correspondance problem in terms of bone length and flexibility, while considering that the skeletal structure is identical.

\section{Related Work}

\subsection{Whole-body imitation}

Indeed, research in whole-body imitation of a human by a humanoid robot have proposed  offline or real-time optimizations to solve inverse kinematics for the end effectors  \cite{6907261,5354271,1242196}. In \cite{6907261} and \cite{1242196}, the authors used mainly inverse kinematics  on each of the kinematic chains for static pose mapping. \cite{5354271} solved the geometric and dynamic differences of the correspondence problem \cite{Nehaniv2002IAA} by scaling by a predefined constant for the difference in length for body parts, and by optimizing inverse kinematics. In summary, they only focused on the end effectors as the key parts to be imitated, and supposed the retargeting is only about the end-effector positions. However, while this might be true for manipulation and task-oriented movements, in other cases, such as for rehabilitation exercises as in \cite{Devanne2018IICRC}, the focus might shift on the intermediate joints. In order to take into consideration all the joints, another method is to use joint orientations as features that should be invariant to the motion performer. However, by simply copying the joint orientations, we  may not properly translate meaningful features of the source motion. For instance, if the source motion contains a contact between the two hands (e.g. clapping hands), and if the target skeleton has shorter forearms than the source skeleton, the hands in the translated motion would not touch. These limitations have encouraged us to explore the use of deep learning methods for motion retargeting.

% A simple approach for motion retargeting consists in computing the joint orientations in the source skeletons and reconstructing the target skeleton combining this orientation information with the known bone lengths of the target skeleton. 

\subsection{Motion retargeting}

As retargeting can be seen as translating a motion from the human embodiment domain to the robot embodiment domain, it entails  extracting the main characteristics from a motion that would be common to human and robot movements, or in other words, finding a common representation between both movements. Domain-to-domain translation is fundamentally a disentanglement problem, where some information content of the source data must be kept while some domain-specific content has to be transformed. For images, typically object positions and features should be kept, while appearance and style is transformed. \cite{Devanne2019CVE2W} proposed shared latent variables as a common representation between movements from a robot and humans with different joint flexibility. However their algorithm does not handle more morphology differences, and needs to be retrained for each exercise : the same model can not be applied to all movements. 
In the case of motion retargeting, we would like to disentangle the motion features from the identity of the subject performing the motion. 
Thus, we hypothesize that a deep learning encoder-decoder architecture may capture such relevant features during the encoding, and properly translate them in the reconstructed motion during the decoding. In order to have a retargeting method that can work with different sources (e.g. different human subjects) and targets (e.g. different robots), we make the encoder invariant with regard to the source skeleton, and the decoder conditioned on the target skeleton. Training such an architecture with paired source and target motions would constitute an ideal supervised learning scenario, but in practice such data is very difficult to collect. One could obtain paired motions in circumstances where different subjects perform the same task (e.g. from action recognition data sets), however it is difficult to ensure that the task is completed with the same motion by different subjects. For instance, for two motions labeled as "picking up a phone", the hand picking up the phone (left or right), the speed of the motion, the initial position of the phone, the pose of the subject (sitting or standing) are features that may not be consistent.

The other option is to work with unpaired data. In this case, we have access to one dataset of motions for the source skeleton and one dataset of motions for the target skeleton, without any known correspondence. Models working with unpaired data have been proposed in the field of image-to-image translation \cite{Zhu2017}, which we try to adapt in this work to perform unpaired cross-domain motion translation between several source and target embodiments.

In order to train a domain-to domain translation without paired data, the CycleGAN method \cite{Zhu2017} leverages a cycle-consistency constraint together with adversarial training. The cycle-consistency criterion ensures that translating back the prediction in the source space yields the initial source image (or motion in our case). The adversarial training constrains the prediction to belong to the target domain. In the UNIT method presented in \cite{Liu2017}, the authors use a similar approach with an encoder-decoder architecture. Using the assumption that both source and target domains share a same latent space, they build a cycle-consistency criterion in the latent space (output of the encoder). Another recent approach \cite{Park2020} replaces this cycle-consistency constraint using contrastive learning techniques. While these methods have been tested successfully for image-to-image translation, our goal is to adapt them to motion retargeting.

In order to use these methods for motion retargeting, we need to build neural network models that can properly process skeleton data. We can take inspiration from the literature in skeleton-based action recognition, which has seen several breakthroughs over the last decade. Initially centered around recurrent and convolutional neural networks \cite{Liu2016, Ke2017}, the field increasingly uses graph convolutions \cite{Yan2018}. Graph convolutions constitute a more natural operation to perform on skeleton data as they allow to properly exploit their underlying structure. Many variants have been proposed, using directed skeleton graphs, using wider kernel size convolutions with $k$-adjacency matrices, or using learnable graph adjacency matrices \cite{Shi2019a, Liu2020}. Performing even better are neural network models using self-attention mechanisms, that can be seen as a special case of graph convolutions where the graph is dynamically constructed based on attention coefficients \cite{Shi2019b, Plizzari2021}.

In the motion retargeting literature, the second step of the human-robot imitation process we described is often considered lightly. Existing approaches typically focus on some joints of interest for which they define a transformation, for instance homothetic \cite{zhang2021human}, or obtained by resizing the bones while keeping the joint angles invariant \cite{choi2019towards}. In comparison, our retargeting method has more freedom in the range of transformations it can apply to the source skeleton.
Most related to our work is the deep motion retargeting model proposed in \cite{Aberman2020}. They also address the problem of unpaired cross-domain translation of motions. They use an architecture consisting of encoders, decoders and discriminators. However, their approach uses as input joint orientations (as quaternions) instead of positions, which has troublesome practical implications for deep learning \cite{Xiang2020}, as well as struggles to capture some meaningful motion features such as joint contacts (e.g. in the clapping hand motion, hand contacts are not reflected in joint orientations information). Moreover, our current raw data are videos, which are processed by deep learning algorithms to extract joint positions \cite{Cao2019ITPAMI,Bazarevsky2020C} with performance comparable to to RGB-D cameras like the Kinect \cite{Marusic2023C2AICHI}, whereas joint orientations are not directly obtained.

\section{Methods}

\begin{figure}[ht]
    \centering
    \vspace{-1em}
    \includegraphics[width=0.9\columnwidth]{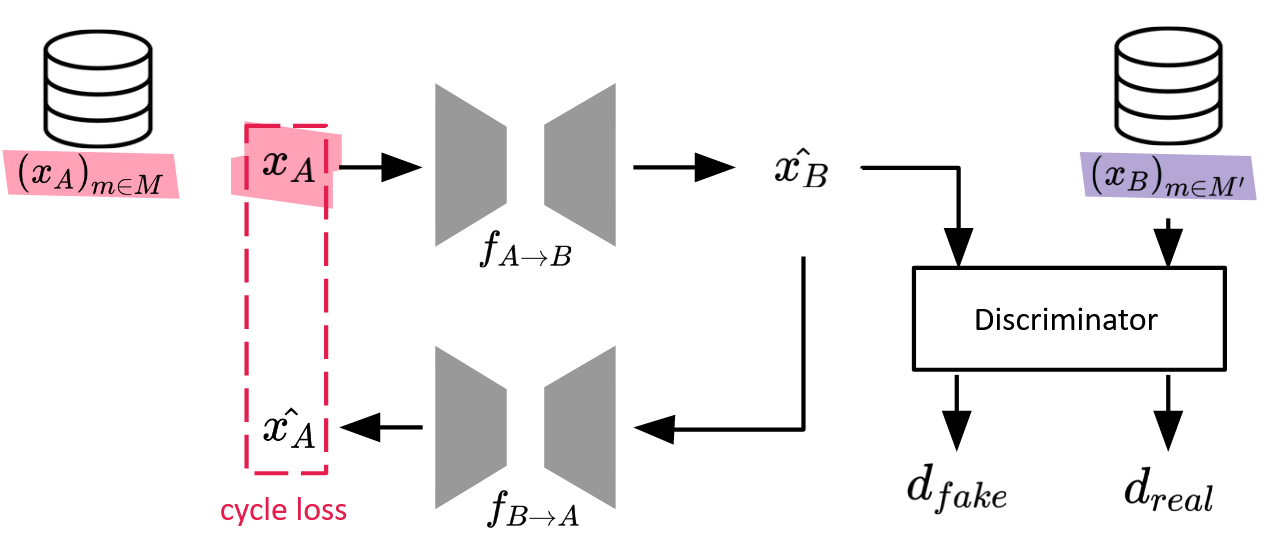}
    \vspace{-1em}
    \caption{Algorithm architecture with an encoder-decoder and a discriminator.}
    \label{fig:archi}
    \vspace{-2em}
\end{figure}

In this section, we describe the retargeting algorithm. As depicted in figure \ref{fig:archi}, the algorithm consists of a pair of encoder-decoder to extract the main characteristics of a motion and to generate a motion in the target joint position space, and a discriminator to challenge the decoder. We then describe our training process.

\subsection{Encoder and decoder}

\begin{figure}[ht]
    \centering
    \vspace{-1em}
    \includegraphics[width=0.9\columnwidth]{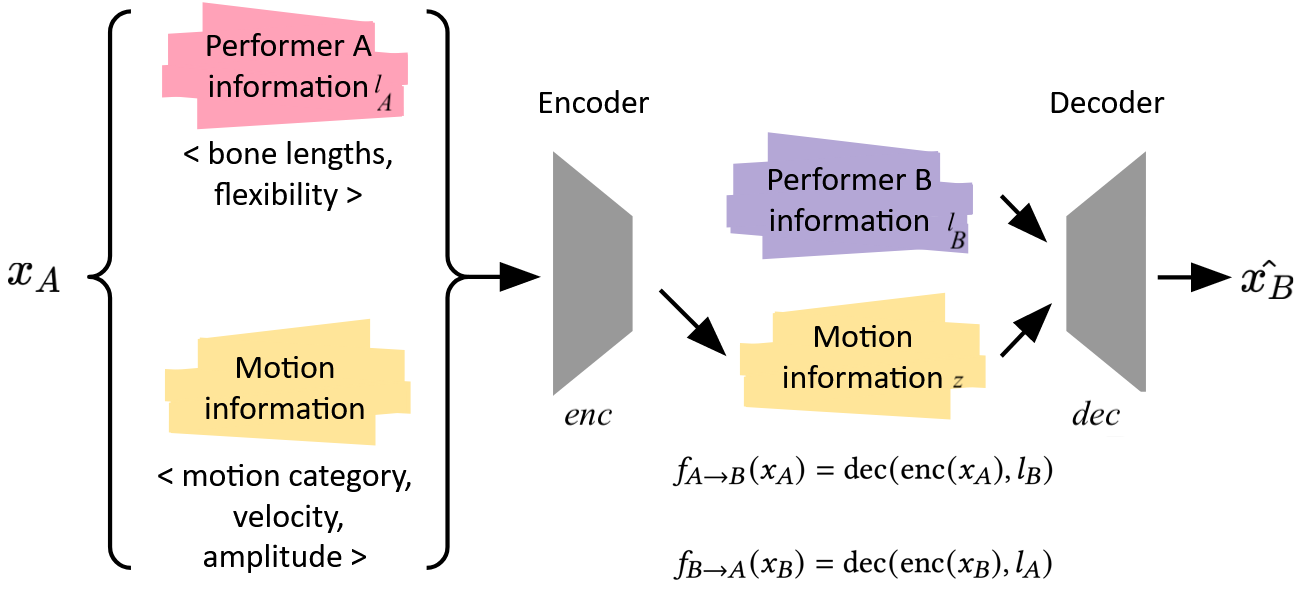}
    \vspace{-1em}
    \caption{Details of the encoder-decoder.}
    \label{fig:encoder_decoder}
    \vspace{-1em}
\end{figure}

We present here the proposed encoder-decoder model for motion retargeting. The encoder takes as input $x_A$, a sequence of joint positions in the domain of motions performed by a performer $A$. It infers a latent variable $z$ from this input motion, disentangling information about the motion itself from the information about the performer, as represented in figure \ref{fig:encoder_decoder}. Then the decoder takes as input this latent variable $z$ as well as the lengths of the performer B bones, that we denote $l_B$, and outputs a prediction $\hat{x_B}$ corresponding to the motion translated to the domain of motions performed by the performer B. In this work, we assume that source and target skeletons share the same structure, but that bone lengths vary between the two.

The encoder and the decoder both comprise three layers of graph convolutions mixed with three layers of temporal 1D convolutions. The graph convolutions allow to process information on the skeleton dimension, while the temporal convolutions allow to process information on the temporal dimension.

For the encoder, the temporal convolutions use a kernel size of 3 and a stride of 2. For the decoder, we experimented with : option 1: transposed convolutions; option 2: using upsampling with a factor of 2, followed by convolutions of kernel size 3 and stride 1. While the transposed convolutions seem at first like the right choice to have a symmetrical architecture for the encoder and the decoder, it can create artifacts on the generated data, that we do not observe using the second option.

For the encoder, we use standard directed graph convolutions, where we have different sets of weights for the parent nodes and for the children nodes (denoted respectively $W_p$ and $W_c$). The graph convolution performs the following operation on an input $x$ of shape ($N$, $d$) where $N$ is the number of nodes in the graph (here the number of joints), and $d$ is the number of features for each node $i$:

\begin{equation}
    x_i \gets \sigma \Big(  W_r \cdot x_i + \sum_{j\in\mathcal{C}(i)} W_c \cdot x_j + \sum_{j\in\mathcal{P}(i)} W_p \cdot x_j \Big)
\end{equation}

\noindent where $\mathcal{P}(i)$ and $\mathcal{C}(i)$ denote respectively the parents and children of node $i$ in the graph, $\sigma$ is the sigmoid activation function, and $W_r$ corresponds to a third set of weights. Bias coefficients are omitted for simplicity, here and in the rest of the methods section.

The decoder needs to include the bone lengths of the target performer as additional input. We choose to include those as edge features in the graph, and instead perform two-steps graph convolutions in the decoder. The first step updates edge features $e_{ij}$ based on adjacent nodes' features $x_i$ and $x_j$, and the second step updates node features based on parent and children edges' features, as expressed in the following equations:

\begin{equation}
    e_{ij} \gets \sigma \Big(  W_{e} \cdot e_{ij} +  W_{-} \cdot x_i  + W_+ \cdot x_j \Big)
\end{equation}

\begin{equation}
    x_i \gets \sigma \Big(  W_r \cdot x_i + \sum_{j\in\mathcal{C}(i)} W_c \cdot e_{ij} + \sum_{j\in\mathcal{P}(i)} W_p \cdot e_{ji} \Big)
\end{equation}

\noindent where $W_e$, $W_+$, $W_-$ denote sets of weights for the first step. This graph convolution operation makes it possible for the decoder to generate motion conditioned on the provided bone lengths.

\subsection{Discriminator}

The discriminator takes as input a predicted motion $\hat{x_B}$ and target performer bone lengths $l_B$ and outputs a real number scoring how well the predicted motion fits the distribution of target motions. It is trained with adversarial training, using positive real samples from the distribution $\mathcal{D}_B$ and fake samples generated by the encoder-decoder network. Its structure is similar to the encoder, with an additional dense layer at the end to output  the score. It combines temporal 1D convolutions and two-steps graph convolutions to include the bone lengths $l_B$ in the processing.

\subsection{Training}

Following the Least Squares GAN method \cite{Mao2017}, the discriminator is trained with the following loss function:

\begin{equation}
    \mathcal{L}_D = \frac{1}{2} \mathbb{E}_{x_B \sim \mathcal{D}_B} \Big[ \big(D(x_B) - 1\big)^2 \Big] + \frac{1}{2} \mathbb{E}_{x_A \sim \mathcal{D}_A} \Big[ D\big(f_{A\rightarrow B}(x_A)\big)^2 \Big]
\end{equation}

\noindent where $D$ denotes the discriminator, and     $ f_{A \rightarrow B}(x_A) = \text{dec}(\text{enc}(x_A), l_B) $  the encoder-decoder network translating from subject A to B.

We train the encoder-decoder network (working as the generator as well) with a combination of several loss functions, following the CycleGAN \cite{Zhu2017} and UNIT \cite{Liu2017} methods for unpaired domain-to-domain translation. Both methods use a loss function coming from the adversarial training:

\begin{equation}
    \mathcal{L}_G = \frac{1}{2} \mathbb{E}_{x_A \sim \mathcal{D}_A} \Big[ \big(D\big(f_{A\rightarrow B}(x_A)\big)-1\big)^2 \Big]
\end{equation}

The cycleGAN model combines it with a cycle consistency loss:
    \vspace{-0.7em}

\begin{equation}
    \mathcal{L}_{cycle} = \mathbb{E}_{x_A \sim \mathcal{D}_A} \Big[ \big\|x_A - f_{B \rightarrow A}(f_{A \rightarrow B}(x_A))\big\|_2^2\Big]
\end{equation}

\noindent where $    f_{B \rightarrow A}(x_B) = \text{dec}(\text{enc}(x_B), l_A) $  denotes the encoder-decoder network translating from subject B to subject A.

In comparison, the UNIT model combines it with a variational auto-encoder loss and a cycle consistency loss on the latent space:

    \vspace{-0.7em}
\begin{equation}
    \mathcal{L}_{vae} = \mathbb{E}_{x_A \sim \mathcal{D}_A} \Big[ \big\|x_A - \text{dec}(\text{enc}(x_A), l_A)\big\|_2^2 + \big\| \text{enc}(x_A) \big\|_2^2\Big]
\end{equation}
\vspace{-2em}

\begin{equation}
    \mathcal{L}_{cycle UNIT} = \mathbb{E}_{x_A \sim \mathcal{D}_A} \Big[ \big\|\text{enc}(x_A) - \text{enc}(f_{A \rightarrow B}(x_A))\big\|_2^2\Big]
\end{equation}

\noindent where enc and dec denote the encoder and decoder networks.

We experiment with both methods, and train our models on a dataset of animated motions called Mixamo (\url{https://www.mixamo.com/}). This dataset advantageously contains the same motions performed by different animated characters. We create a training and testing set from motions exported from the website, with 800 motions distributed among 25 characters for the training set (unpaired data), and 110 same motions for 4 other characters in the test set. The Mixamo dataset comprise fbx files, a type of 3D model file containing mesh, material, texture, and skeletal animation data, which can be easily 3d joint positions.

With this separation, we ensure that the training data do not contain any motion performed by two different characters, and on the contrary, that testing data contain corresponding motions for different characters, which allows us to compute a prediction error. However, the motions are automatically adapted to the animated characters and  the implementation of this adaptation is not available. We can see that some motions present impossible body configurations (e.g. arm going through head). Consequently, we will also perform visual inspection in order to validate our method.

\section{Early results}

\begin{figure}[h]
    \centering
    \vspace{-0.7em}
    \includegraphics[width=\columnwidth]{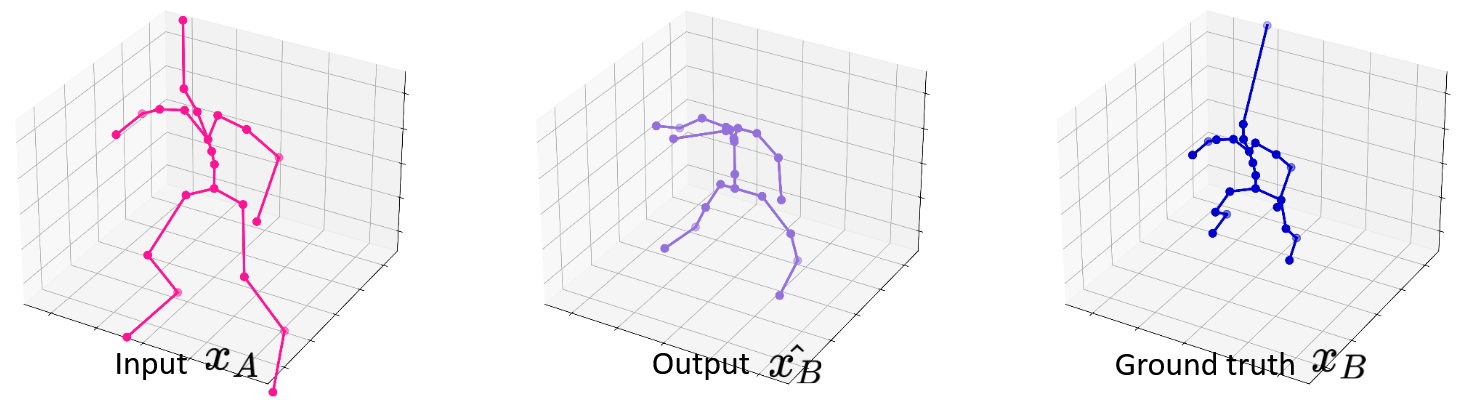}
    \vspace{-1.5em}
    \caption{Example of motion retargeting using UNIT.}
    \vspace{-0.8em}
    \label{fig:example_unit}
\end{figure}

We present here early results. Figure \ref{fig:example_unit} displays an example motion $x_A$, the corresponding predicted motion obtained with the UNIT model after training $\hat{x_B}$, and the ground truth motion performed by performer B taken from the Mixamo test set $x_B$. Figure \ref{fig:example_unit} only shows one time frame of the full motion.

While we can observe that the predicted skeleton is of comparable size with the ground truth skeleton (which indicates that there is to some extent a translation of the motion to the target domain), the pose is visually very different from the ground truth.

We measure the average retargeting prediction error on the test set using the two methods described above, and compare them with two simple baselines consisting respectively in copying the source joint positions, and copying the source joint orientations (and computing the joint positions using the target skeleton bone lengths). The results are displayed in table \ref{tab:results}.

\begin{table}[h]
  \vspace{-1em}
  \caption{Comparison of the different retargeting methods.}
  \label{tab:results}
    \vspace{-1em}
  \begin{tabular}{cccc}
    \toprule
    Method & \makecell{Reconstruction \\error (train)} & \makecell{Reconstruction \\ error (test)} & \makecell{Retargeting \\ error (test)} \\
    \midrule
    \makecell{Position copy} & 0 mm & 0 mm & 195 mm \\
    \makecell{Rotation copy} & 0 mm & 0 mm & \textbf{79 mm} \\
    CycleGAN & 70 mm & 182 mm & 243 mm \\
    UNIT & 48 mm & 164 mm & 209 mm \\
  \bottomrule
    \vspace{-2.8em}
\end{tabular}
\end{table}

The results argue so far against our initial hypothesis that deep learning methods can successfully perform unpaired motion retargeting, as a simple method copying joint orientations (rotation copy in the table) can better retarget motions. 

More experiments will be conducted in order to properly identify the causes of this failure. While unsupervised domain-to-domain translation has been successfully applied to images, the mechanisms behind this success remain unclear. Indeed, the unsupervised domain-to-domain translation constitutes an ill-posed problem, where many solutions can satisfy the criteria we optimize for, while not performing the translation we expect. For instance, an image-to-image style translation problem in an unsupervised setting could in principle accept as a solution a function $f_{A\rightarrow B}$ correctly applying the desired style translation $f^*_{A\rightarrow B}$, but also applying at the same time a bijection $b$ on the image space for which the style is invariant. The candidate solutions $f_{A\rightarrow B} = f^*_{A\rightarrow B} \circ b$ and $f^*_{A\rightarrow B}$ are as good at minimizing the loss function, yet only the later is the solution we wish to reach. Investigating this problem, and how it was addressed for image-to-image translation, could help us find better network designs or initialization strategies leading to the desired solution.

\section{Conclusion}

This early-stage research has shown that deep learning unsupervised motion retargeting is feasible yet not accurate enough to replace simpler naive methods. Still, as explained in the introduction, such naive methods are limited in that they cannot capture and translate some meaningful features of the motion, and we think that more effort is needed to improve motion retargeting models. 

Future work will extend the current study in three directions:
\begin{itemize}
    \item Further investigating the failure of the current method, as explained in the last section.
    \item Creating a dataset of paired motion data from human-human imitation or robot-human imitation. We hypothesize that humans can perform accurate imitation, and thus participate in building a dataset of paired motions. A first step would be to have enough paired data to replace the Mixamo test set. In a second step, if enough paired data is available, it might become possible to train the retargeting models in a supervised learning setting, largely simplifying the problem.
    \item Improving the model architecture in order to obtain more accurate retargeting predictions. As hinted by their success in the field of action recognition, skeleton self-attention layers, as well as temporal self-attention layers (transformers) could also help our model better capturing important features of the motion and generating accurate predictions.
\end{itemize}

\begin{acks}
This project is partially funded by Institut Carnot and AID Project ACoCaTherm.
\end{acks}

\bibliographystyle{ACM-Reference-Format}
\balance
\bibliography{references}

%\appendix

\end{document}